\pdfoutput=1

\documentclass[11pt]{article}

\usepackage{ACL2023}

\usepackage{times}
\usepackage{latexsym}
\usepackage{color,soul}
\usepackage{multirow}
\usepackage[T1]{fontenc}

\usepackage[utf8]{inputenc}

\usepackage{microtype}

\usepackage{inconsolata}

\title{Factuality Detection using Machine Translation - a Use Case for German Clinical Text}

\author{Mohammed Bin Sumait, Aleksandra Gabryszak, Leonhard Hennig \and Roland Roller \\ German Research Center for Artificial Intelligence (DFKI)\\
Speech and Language Technology Lab \\
\texttt{\{firstname.lastname\}@dfki.de}}

\begin{document}
\maketitle
\begin{abstract}
\vspace{-0.3cm}
Factuality can play an important role when automatically processing clinical text, as it makes a difference if particular symptoms are explicitly not present, possibly present, not mentioned, or affirmed. In most cases, a sufficient number of examples is necessary to handle such phenomena in a supervised machine learning setting. However, as clinical text might contain sensitive information, data cannot be easily shared. In the context of factuality detection, this work presents a simple solution using machine translation to translate English data to German to train a transformer-based factuality detection model. 
\end{abstract}
\vspace{-0.5cm}
\section{Introduction}

Factuality refers to the concept that a speaker can present statements about world events with varying degrees of uncertainty as to whether they happened. Factuality reflects, for instance, if an event is affirmed, negated, or uncertain. In the medical domain, detecting if symptoms or diseases are signaled as present, not present, possibly or doubtfully present, and therefore uncertain is essential. 
Detecting factuality is challenging since it can be expressed by very different linguistic categories (e.g. verbs, nouns, adjectives, adverbs), plus it must be taken into account how they are embedded in a sentence \cite{RudingerEtAlFactuality2018}.
Additionally, linguistic factuality cues can be very domain-specific, so the availability of relevant datasets is essential.

Classical supervised machine learning requires training data, and, at the same time, most existing datasets are published in English. In addition, clinical text contains sensitive patient data, which often makes it difficult to share due to ethical and legal aspects. Although the situation has slowly changed regarding the availability of German clinical text resources \cite{rohrig2022grascco}, many other languages suffer a similar situation. Conversely, the quality of machine translation has significantly improved in the last decade, also regarding the translation of biomedical text/publications, including clinical case reports \cite{neves-etal-2022-findings}. For this reason, this work explores the usage of machine translation to create (translated) text resources for factuality detection in German clinical text. 

Clinical notes are short text documents written by physicians during or shortly after the treatment of a patient. In general, this kind of text contains much valuable information about the current health condition, as well as treatment, of the patient. They differ from biomedical publications and clinical case reports, as notes are often written under time pressure with a high information density, a telegraphic writing style, non-standardized abbreviations, colloquial errors, and misspellings. Therefore, it is unclear if current machine translation systems can handle this text, considering that data might contain sensitive information and should not be shared with a third party outside the hospital. %

This work makes the following contributions: 1) We successfully use a local machine translation to train a model for factuality detection on German clinical text. 2) Our model %
outperforms the only `competitor' NegEx, and 3) will be published as open access model\footnote{https://huggingface.co/binsumait/factual-med-bert-de}. Finally, 4) for those interested in NegEx, we release it as a modular PyPI package with a few important fixes\footnote{https://github.com/DFKI-NLP/pynegex} and also propose improvement suggestions to the used trigger sets.

\vspace{-0.3cm}
\section{Methods and Data}

The idea of this work is based on the usage of machine translation to generate a German corpus to train a classifier dealing with factuality in clinical text. In the following, we outline the approach, the necessary methods, and the dataset used. %

\begin{table*}[htbp!]
    \centering
    \small
    \begin{tabular}{c|p{5.7cm}|p{7.6cm}}
    \textbf{Factuality} & \textbf{English} & \textbf{German translation} \\
    \hline
    affirmed & Clinically, a <E>severe neuropsychological syndrome</E> was found when the patient was taken over. & Klinisch fand sich bei Übernahme des Patienten in <E>schweres neuropsychologisches Syndrom</E>. \\ \hline
    negation & Patient denies <E>headache</E>. & Patient verneint <E>Kopfschmerzen</E>.  \\ \hline
    possible & Thus, a <E>tumour</E> cannot be ruled out. & Ein <E>Tumor</E> kann daher nicht ausgeschlossen werden. \\
    \hline
    \end{tabular}
    \caption{Example sentences with target entities, factuality label, and possible translations.}
    \vspace{-0.35cm}
    \label{tab:translation_ex}    
\end{table*}

\subsection{Factuality Detection}

In literature, (medical) factuality detection is often reduced to a simple classification. Given a sentence and an entity, the task is to define the factuality of the entity in the given context. In most cases, the entity of interest is a symptom or medical condition. Most related work targets the three classes \textbf{affirmed}, \textbf{negated} and \textbf{possible}. However, as simple as this sounds, factuality cannot always be easily mapped to those few classes. 

One of the most prominent tools to deal with factuality in the medical text is NegEx \cite{chapman2001}, a rule-based approach with pre-defined regular expressions, so-called triggers, and can detect the three aforementioned factuality classes. It achieves, particularly in the context of negations, quite good results on clinical text. Hedges instead offer more possibilities for how they are described, therefore achieving a much lower performance. Initially, it was developed for English, but over the years, it has also been translated into other languages, such as Spanish or Swedish \cite{cotik2016syntactic,chapman2013extending}. In addition, many alternative (machine learning) solutions have been published in the last two decades. We refer to the overview by \citet{khandelwal2019} for more details. For German, however, only one negation detection exists, which relies on the NegEx solution and uses a set of translated trigger words (English to German) \cite{cotik2016negation}.

\subsection{Data}

In the following, we briefly introduce the data used for this work. First, we present i2b2, which has been used for machine translation and to train our model. In addition, we later test our model on additional German data, namely Ex4CDS and NegEx-Ger, and in the appendix also BRONCO150. 

The \textbf{2010 i2b2/VA} data \cite{uzuner2010} consists of English medical text and includes three tasks - extraction of concepts, assertions identification, and relation detection. In this work, we focus on the assertion task. Overall a total of six assertion types were considered, namely present, absent, possible, conditional, hypothetical and not associated with the patient. However, this work focused only on the first three labels, as only those are considered within NegEx. i2b2 data is translated to German to train a German machine learning model.

\textbf{Ex4CDS} \cite{roller2022annotated} is a small dataset of physicians' notes containing explanations in the context of clinical decision support. The notes are written in German and include various annotation layers, including factuality. As the data includes multiple factuality labels, we reduced the labels to our three target labels, mapping \textit{possible-future} and \textit{unlikely} to \textit{possible}, and \textit{minor} to \textit{affirmed}. As target entities, we consider only sentences containing \textit{medical-conditions}.

\textbf{NegEx-Ger} is a small dataset consisting of sentences taken from clinical notes and discharge summaries and has been used initially to evaluate the German NegEx version in \citet{cotik2016negation}. For our use case, the data has been used for testing, and for this, we merged the sentences of both clinical text types. However, the number of sentences containing the possible label is small (22 for discharge summaries and 4 for clinical notes).

\subsection{Translation Approach}

For our proposed idea, two aspects need to be considered: First, we aim at a solution that could be applied to sensitive data. Therefore, the machine translation component must run locally. This means we cannot rely on the variety of existing state-of-the-art online approaches. Second, as we define factuality as a classification problem with a given sentence (context) and an entity, our translations need to keep track of the target entity within a sentence. A simple example is given in Table~\ref{tab:translation_ex}, which shows an English sentence with a target entity `headache' and the label `negation'. The German translation needs to keep the focus on the target entity.

In this work, we rely on TransIns \cite{steffen2021}, an open-source machine translation that can be installed locally. TransIns is built on MarianNMT \cite{junczys2018} framework and enables translating texts with an embedded markup language. Specifically, we translate sentences with tagged entities, as shown in Table \ref{tab:translation_ex}.

A manual inspection revealed multiple problems with the translations: In some cases (roughly 40\% of the issues), translations were corrupt as they contained cryptic and/or repetitive text sequences that were foreign from the original text. Such noise patterns could partially or entirely affect the target texts' context. Or, in very few cases (only 4\%), no translation output could be produced. In the rest of the cases, the markup no longer included the target entity. In any way, such output has been discarded from the data, and we resulted in 18,297 data points (initially 18,397), which we used to train and evaluate our machine learning model.

\section{Experiments and Results}
\label{sec:experiments-results}

We conduct three different experiments - starting with the English i2b2 data, we use Bio+Discharge Summary BERT \cite{alsentzer2019} and compare the results to NegEx. Similar experiments have also been conducted in other papers. However, in our case, those results serve as a comparison. Thus, the model is not optimized to achieve the best possible performance. Next, we train German-MedBERT \cite{shrestha2021} on the translated i2b2 data and compare the results to the performance of the German NegEx implementation. Finally, we apply both German factuality approaches to different German medical texts to determine how well the models perform in a more realistic setup.

\begin{table}[htbp!]
    \centering
    \small
    \begin{tabular}{l|ccc|ccc}
    & \multicolumn{3}{c|}{NegEx}  & \multicolumn{3}{c}{BERT-based}  \\\hline
    Label  & Prec  & Rec & F1 & Prec & Rec & F1  \\ \hline
    E Affirmed & 0.88 & 0.97 & 0.93 & \textbf{0.97}& \textbf{0.99} & \textbf{0.98}  \\
    N Negated & 0.89 & 0.79 & 0.84 & \textbf{0.98} & \textbf{0.97} & \textbf{0.97}  \\ 
    G Possible & 0.79 & 0.04 & 0.08 & \textbf{0.85} & \textbf{0.64} &  \textbf{0.73} \\ \hline
    G Affirmed & 0.84 & 0.96 & 0.90 & \textbf{0.96} & \textbf{0.98} & \textbf{0.97}   \\
    E Negated & 0.83 & 0.65 & 0.73 & \textbf{0.95} & \textbf{0.93} & \textbf{0.94}   \\
    R Possible & 0.28 & 0.02 & 0.04 & \textbf{0.80} & \textbf{0.64} & \textbf{0.71}   \\
    \hline
    \end{tabular}
    \caption{Performance results between NegEx baselines and BERT-based models on the original English i2b2 dataset (upper part) and German translation (lower part).}
    \label{tab:i2b2_results}    
\end{table}

The results of the first two experiments are presented in Table \ref{tab:i2b2_results} and show various interesting findings: Firstly, NegEx provides impressive results on the affirmed label, good results for negations, and unsatisfying results for the possible label. Moreover, on both datasets, English and German, the BERT-based model outperforms NegEx, on all scores. Additionally, results on the English dataset are always higher than those on the translated dataset. This might be unsurprising as data quality decreases. %
Finally, the table shows that BERT-based models show a substantial increase in performance for the possible label.

\begin{table}[htbp!]
    \centering
    \small
    \begin{tabular}{l|ccc|ccc}
     & \multicolumn{3}{c|}{NegEx}  & \multicolumn{3}{c}{BERT-based}  \\\hline
    Label  & Prec  & Rec & F1 & Prec & Rec & F1  \\ \hline
    N Affirmed & 0.96 & 0.94 & 0.95 & \textbf{0.97} & \textbf{0.96} & \textbf{0.96}  \\
    E Negated & 0.93 & 0.96 & 0.95 & \textbf{0.97} & \textbf{0.98} & \textbf{0.97}  \\
    G Possible & 0.46 & 0.50 & 0.48 & \textbf{0.50} & 0.50 & \textbf{0.50}  \\ \hline
    E Affirmed & 0.85 & 0.88 & 0.86 & \textbf{0.88} & \textbf{0.92 }& \textbf{0.90}  \\
    X Negated & 0.66 & 0.89 & 0.76 & \textbf{0.86} & \textbf{0.95} &  \textbf{0.90} \\
    4 Possible & 0.50 & 0.18 & 0.26 & \textbf{0.61} & \textbf{0.38} & \textbf{0.47}  \\ \hline
    \end{tabular}
       \caption{Performance results on different German medical text sources, namely the original German NegEx (upper part), and Ex4CDS dataset (lower part).}
    \label{tab:applied_to_German}
\end{table}

Table \ref{tab:applied_to_German} presents the performance of the NegEx and the BERT-based model on two German datasets. In the upper part of the table, the results on NegEx-Ger are presented and the results on Ex4CDS are in the lower part. Similarly, as on the translated i2b2 dataset in Table \ref{tab:i2b2_results}, the machine learning model outperforms NegEx. However, this time the performance gain is not so strong anymore. The NegEx-Ger is small and relatively homogeneous (regarding the variety of negations), and NegEx already performs well on the negations. Therefore the machine learning model achieves only a performance boost of two points in F1. In case of possible, the number of examples might be too small to see the benefit of the ML model.

On Ex4CDS data, NegEx already struggles with \textit{negated} (0.76) and performs low in the case of \textit{possible} (0.26) - although the results are much better in comparison to the results on i2b2 (English and German). Here, the machine learning model leads to a performance boost of 14 points for \textit{negated} and 21 points for \textit{possible}. 

\section{Analysis and Discussion}

Our results indicate that we can successfully apply machine translation to generate a German clinical dataset to train a machine learning model with. Most notably, this model can outperform NegEx, which partially already provides satisfying results. While it is important that  a negation detection tool for German clinical text needs to run within a hospital infrastructure, it might be questionable if BERT-based approaches might be the right solution, as it requires much more hardware resources than the simple NegEx solution. This is supported by the results on NegEx-Ger, in which the BERT achieves only a minor performance gain. However, as this data is small and homogeneous, the results on Ex4CDS affirm the usage of machine learning, as we achieve a notable performance gain. Note, information about the frequency of each label in the test data is provided in the appendix. As our BERT model was trained on potential suboptimal translations, we analyse some errors in more detail in the following.

\subsection{Linguistic Error Analysis}
\label{sec:linguistic-analysis}

Our analysis focuses on the prediction errors caused by the translation or by differences in the features of the German and English language.
Table~\ref{tab:error-cause} contains full-text examples illustrating the issues described below.

In various cases, a factuality cue was completely missing in the translation, or the sense of the cue was not preserved (e.g., \textit{to rule out} was translated with \textit{Vorschriften} instead of \textit{ausschließen}%
). In those cases, NegEx and BERT labeled the instances wrongly as affirmations.

In other cases, we observe that the factuality cues are outside of the original data's entities but in the translation they are placed within the entity markup. That is often correlated with the prediction changing from negation or possible to affirmation. For example, both NegEx and BERT correctly recognized the negated assertion of the original phrase \textit{did not notice [any blood]}, whereas both German models consider the translation \textit{bemerkte [kein Blut]} as affirmed in which the negation cue (\textit{not} / \textit{kein}) became part of the entity.

For NegEx, a further problem are missing factuality cues in the trigger list. For example, it systematically does not recognize the cue \textit{verleugnen} (one of the possible translations of the word \textit{deny}, which is included in the English NegEx). %
Additionally, some problems with factuality cues are specific to the German language and require additional handling:
(a) German compounds must be written as one word; unfortunately, German NegEx cannot handle cases when a compound consists of words referring to a medical problem and its negation (e.g. \textit{schmerzfrei} / \textit{pain free}), since it seems not to recognize a  factuality cue if it is not written as a separate phrase,
(b) cues with umlauts in text such as \textit{aufgelöst} seem not to be recognized, because the umlauts are encoded as \textit{oe} in the German trigger list,
(c) missing possible word orders of factuality phrases (e.g. word order might depend on the embedding syntactic structure; e.g. \textit{wurde ausgeschlossen} vs. \textit{ausgeschlossen wurde} in a main vs. subordinate clause).

\section{Related Work}
\noindent\textbf{Machine Translation for Cross-lingual Learning}
MT is a popular approach to address the lack of data in cross-lingual learning~\cite{pmlr-v119-hu20b,yarmohammadi-etal-2021-everything}. There are two basic options - translating target language data to a well-resourced source language at inference time and applying a model trained in the source language~\cite{Asai2018MultilingualER,cui-etal-2019-cross}, or translating source language training data to the target language, while also projecting any annotations required for training, and then training a model in the target language~\cite{khalil-etal-2019-cross,kolluru-etal-2022-alignment,frei_2023}. Both approaches depend on the quality of the MT system, with translated data potentially suffering from translation or alignment errors~\cite{aminian-etal-2017-transferring,ozaki-etal-2021-project}. While the quality of machine translation for health-related texts has significantly improved~\cite{neves-etal-2022-findings}, using MT in the clinical domain remains underexplored, with very few exceptions~\cite{frei_2023}.

\noindent\textbf{Factuality Detection}
Previous research focused mainly on assigning factuality values to events and often framed this task as a multiclass classification problem over a fixed set of uncertainty categories~\cite{rudinger-etal-2018-neural-models,zerva2019automatic,pouran-ben-veyseh-etal-2019-graph,qian-etal-2019-document,bijl-de-vroe-etal-2021-modality,vasilakes-etal-2022-learning}. 
In the biomedical/clinical domain, \citet{uzuner2010} present the i2b2 dataset for assertion classification, and \citet{Thompson2011EnrichingAB} introduce the Genia-MK corpus, where biomedical relations have been annotated with uncertainty values. \citet{vanAken2021AssertionDI} release factuality annotation of 5000 data points sourced from MIMIC. %
 \citet{Kilicoglu2017AssigningFV} introduce a dataset of PubMed abstracts %
 with seven factuality values, and find that a rule-based model is more effective than a supervised machine learning model on this dataset.

\section{Conclusion}

This work presented a machine learning-based factuality detection for German clinical text. The model was trained on translated i2b2 data and tested, first on the translations and then on other German datasets and outperformed an existing method for German, NegEx. The simple machine translation approach might interest the Non-English clinical text processing community. The model will be made publicly available.%

\section*{Ethical Considerations}
We use the original datasets “as is”. Our translations of i2b2 thus reflect any biases of the original dataset and its construction process, as
well as biases of the MT models (e.g., rendering gender-neutral English nouns to gendered nouns in German). We use BERT-based PLMs in our experiments, which were pretrained on a large variety of medical source data. Our models may have inherited biases from these pretraining corpora.

Since medical data is highly sensitive with respect to patient-related information, all datasets used in our work are anonymized. The authors of the original datasets~\cite{uzuner2010,roller2022annotated} have stated various measures that prevent collecting sensitive, patient-related data. Therefore, we rule out the possible risk of sensitive content in the data.

\section*{Limitations}
A key limitation of this work is the dependence on a machine translation system to get high-quality translations and annotation projections of the source language dataset. Depending on the availability of language resources and the quality of the MT model, the translations we use for training and evaluation may be inaccurate, or be affected by translation noise, possibly leading to overly optimistic estimates of model performance. In addition, since the annotation projection is completely automatic, any alignment errors of the MT system will yield inaccurate instances in the target language.

\section*{Acknowledgements}
This research was supported by the German Research Foundation (DFG, Deutsche Forschungsgemeinschaft) through the project KEEPHA (442445488) and the German Federal Ministry of Education and Research (BMBF) through the projects KIBATIN (16SV9040) and CORA4NLP (01IW20010).

\bibliography{main}
\bibliographystyle{acl_natbib}

\appendix

\section{Appendix}
\label{sec:appendix}

The main contribution of this short paper was to show that it is possible to develop a machine learning-based factuality detection for non-English, without training examples in the given language - just by using a local machine translation. In addition, we would like to present a small `bonus' experiment, which did not fit into the main article anymore. More precisely, we wanted to find out how the performance of such a model changes if data in a reasonable size is available for training. The additional experiment is presented in Appendix A.1, followed by some additional text examples for the linguistic error analysis and some further information.

\subsection{Additional Experiment}

The additional experiment has been conducted with the \textbf{BRONCO150} \cite{kittner2021annotation} dataset, a relatively large corpus originating from 150 German oncological de-identified discharge summaries and annotated for multiple tasks, including factuality detection. For our experiment, we consider only the target entities \textit{diagnosis}. Similar to Ex4CDS, it has various factuality values, which we mapped to our three target labels, namely \textit{possible future} and \textit{speculation} to \textit{possible}. Note, BRONCO150 contains various fragmented entities (entities split into two to three parts). For our experimental setup, we merged entity fragments and considered only those sentences with not more than 50 characters between the fragments.

The label distribution of the obtained BRONCO150 data and the distribution of the other datasets from the main paper are presented in Table \ref{tab:labels_distribution}.

First, we run the same experiment as presented in Table \ref{tab:applied_to_German}, also on BRONCO150 data. The results using our FactualMedBERT-DE model are presented in Table \ref{tab:applied_to_German_full}.

\begin{table}[htbp!]
    \centering
    \small
    \begin{tabular}{l|ccc|ccc}
     & \multicolumn{3}{c|}{NegEx}  & \multicolumn{3}{c}{BERT-based}  \\\hline
    Label  & Prec  & Rec & F1 & Prec & Rec & F1  \\ \hline
    N Affirmed & 0.96 & 0.94 & 0.95 & \textbf{0.97} & \textbf{0.96} & \textbf{0.96}  \\
    E Negated & 0.93 & 0.96 & 0.95 & \textbf{0.97} & \textbf{0.98} & \textbf{0.97}  \\
    G Possible & 0.46 & 0.50 & 0.48 & \textbf{0.50} & 0.50 & \textbf{0.50}  \\ \hline
    E Affirmed & 0.85 & 0.88 & 0.86 & \textbf{0.88} & \textbf{0.92 }& \textbf{0.90}  \\
    X Negated & 0.66 & 0.89 & 0.76 & \textbf{0.86} & \textbf{0.95} &  \textbf{0.90} \\
    4 Possible & 0.50 & 0.18 & 0.26 & \textbf{0.61} & \textbf{0.38} & \textbf{0.47}  \\ \hline
   B Affirmed & 0.87 & 0.96  & 0.91  & \textbf{0.88} & \textbf{0.97} & \textbf{0.92}   \\
   R Negated & 0.69 & 0.66 & 0.68 & \textbf{0.75}  & \textbf{0.80} & \textbf{0.77}   \\
   O Possible &  0.68 & 0.24 & 0.36 & \textbf{0.73} & \textbf{0.25} & \textbf{0.37}  \\
    \hline
    \end{tabular}
       \caption{Performance results on different German medical text sources, namely the original German NegEx (upper part), the Ex4CDS dataset (middle) and BRONCO150 (lower part).}
    \label{tab:applied_to_German_full}
\end{table}

\begin{table}[htbp!]
    \centering
    \small
    \begin{tabular}{l|ccc}
          & Affirmed & Negated & Possible   \\ \hline
     2010 i2b2/VA & 7603 & 2305 & 595   \\
    Ex4CDS & 892 & 225 & 179  \\
    NegEx-Ger & 645 & 443 & 26   \\ 
    BRONCO150 & 3179 & 331 & 523 \\ \hline
    \end{tabular}
    \caption{Support numbers in the evaluation sets for each processed dataset.}
    \label{tab:labels_distribution}
\end{table}

Next, we train two additional models, one on a BRONCO150 training split and a second using the BRONCO150 train together with the translated i2b2 data. Both models were initialized from the same model as that of FactualMedBERT-DE.
Table~\ref{tab:bert_models} compares our FactualMedBERT-DE against the other two BERT-based models on the different datasets. %

\begin{table*}[!htb]
    \centering
    \small
    \begin{tabular}{p{2.5cm}|l|ccc|ccc|ccc|ccc}
     \multicolumn{2}{l|}{}& \multicolumn{3}{c|}{2010 i2b2/VA}  & \multicolumn{3}{c|}{NegEx-Ger}  & \multicolumn{3}{c|}{Ex4CDS}  & \multicolumn{3}{c}{BRONCO150}  \\ \hline
    Model & Label  & Prec  & Rec & F1 & Prec & Rec & F1  & Prec  & Rec & F1  & Prec  & Rec & F1 \\ \hline
   \multirow{3}{2.5cm}{FactualMedBERT-DE} & Affirmed & 0.96 & 0.98 & \textbf{0.97} & 0.97 & 0.96 & 0.96 & 0.88 & 0.92 & 0.90 & 0.88 & 0.97 & 0.92 \\ & Negated & 0.95 & 0.93 & \textbf{0.94} & 0.97 & 0.98 & 0.97 & 0.86 & 0.95 & 0.90 & 0.76 & 0.79 & 0.78  \\  & Possible & 0.80 & 0.64 & \textbf{0.71} & 0.50 & 0.50 & 0.50 & 0.61 & 0.38 & 0.47 & 0.68 & 0.19 & 0.30  \\ \hline
   
       \multirow{3}{2.5cm}{BRONCO150-BERT} & Affirmed &  0.88 & 0.95 & 0.92  & 0.97 & 0.92 & 0.94 & 0.90 & 0.90 & 0.90 & 0.96     & 0.96 & 0.96 \\ & Negated & 0.95 & 0.67 & 0.79 & 0.97 & 0.97 & 0.97 & 0.89 & 0.88 & 0.88 & 0.95 & 0.83 & \textbf{0.89} \\  & Possible & 0.42 & 0.47 & 0.44 & 0.28 & 0.65 & 0.39 & 0.56 & 0.59 & 0.58 & 0.76 & 0.84 & \textbf{0.80}  \\ \hline
       
       \multirow{3}{2.5cm}{i2b2+BRONCO150 BERT} & Affirmed & 0.94 & 0.98 & 0.96 & 0.98 & 0.95 & 0.96 & 0.90 & 0.94 & \textbf{0.92} & 0.95 & 0.98 & \textbf{0.97} \\ & Negated & 0.96 & 0.91 & 0.93 & 0.98 & 0.97 & 0.97 & 0.90 & 0.91 &   \textbf{0.91} & 0.93 & 0.83 & 0.88 \\ & Possible & 0.82 & 0.54 & 0.65  & 0.39 & 0.73 & \textbf{0.51} & 0.70 & 0.54 & \textbf{0.61} & 0.85 & 0.74 & 0.79  \\ 
    \hline

    \end{tabular}
       \caption{Performance results of three BERT models trained on translated i2b2 (FactualMedBERT-DE), BRONCO150 and 2010 i2b2 + BRONCO150, respectively. The models were evaluated on different German medical text sources, namely our translated i2b2 2010 test set, the German NegEx, the Ex4CDS dataset and BRONCO150 test set. For each dataset, best per-label F1-performances are displayed in \textbf{bold}.}
    \label{tab:bert_models}
\end{table*}

\begin{table*}[!htb]
    \centering
    \small
    \begin{tabular}{p{4.1cm}|p{5.3cm}|p{5.3cm}}
    
    Issue & English & German \\\hline
    
    missing trigger in translation & The patient radiated down her left arm associated with some nausea, \underline{no} <E> shortness of breath </E>, cough, vomiting, diarrhea. & \textit{Die Patientin strahlte in Verbindung mit Übelkeit, <E> Atemnot, </E> Husten, Erbrechen, Durchfall nach unten.}\\\hline
    incorrect trigger translation & 
    \textit{\underline{RULE OUT FOR} <E> myocardial infarction </E>} & \textit{\underline{VORSCHRIFTEN FÜR} <E> den Myokardinfarkt </E>}  \\\hline

    trigger in the translation is within the entity & \textit{She did \underline{not} notice <E> any blood / urine / emesis / stool in the bed </E>}. & Sie bemerkte <E> \underline{kein} Blut / Urin / Erbrechen / Stuhl im Bett. </E> \\\hline
    
    missing of a possible trigger translation in NegEx-Ger & \textit{\underline{Denies} <E> fevers </E>, pleuritic chest pain or cough.} & \textit{\underline{Verleugnet} <E> Fieber, </E> pleuritische Brustschmerzen oder Husten.}\\\hline
    
    missing of translated compounds of type Entity + trigger in NegEx-Ger & \textit{She was <E> pain </E> \underline{free} on the day of discharge .} & \textit{Sie war am Tag der Entlassung <E> schmerz\underline{frei}. </E>}\\\hline
    
    missing trigger phrase in NegEx-Ger due to word order & \textit{He then presented to Mass. Mental Health Center where he \underline{ruled out for} <E> an myocardial infarction </E> by enzymes and electrocardiograms.} & \textit{Er überreichte dann der Messe. Mental Health Center, wo er für <E> einen Myokardinfarkt </E> durch Enzyme und Elektrokardiogramme \underline{ausgeschlossen wurde}.} \\\hline
    different encoding of umlauts in text and NegEx-Ger & \textit{<E>the hypernatremia</E> fully \underline{resolved} when he resumed eating on his own and had access to free water .
} & \textit{<E>Die Hypernatrimie</E> vollständig \underline{aufgeloest}, als er wieder essen auf eigene Faust und hatte Zugang zu freien Wasser.}\\\hline

    \hline
    \end{tabular}
    \caption{Examples of the potential causes for prediction errors. The analysis focuses on the translation problems and  the differences between the German and English language. The tags <E></E> enclose the entities, the factuality triggers are underlined. The original English examples originate from the i2b2 data.}
    \label{tab:error-cause}
\end{table*}

\paragraph{Brief discussion:} The results show that each model performs best on the data of the same dataset - FactualMedBERT-DE on the translated i2b2 data and BRONCO150-BERT on the BRONCO150 data - this is no surprise. Moreover, the results indicate that the mixed model (i2b2+BRONCO150-BERT) performs generally well on all datasets, therefore might be the model of choice. However, it is important to note, that BRONCO150 has got an unusual label distribution. While \textit{affirmed} is the most frequent label in all datasets, BRONCO has got an unusually high frequency of \textit{possible} labels, which is connected to the way labels were mapped to the three final actuality labels. However, this might influence the actuality classification of other datasets.

\subsection{BERT Setup}
For BERT, we used epochs number of 3/4 (for English and German BERT, respectively), a batch size of 32, a dropout rate of 0.1, and a learning rate of $1e-5$.

\subsection{Examples of Linguistic Error Analysis}

Our analysis focuses on the potential sources for false predictions, in particular on causes related to the translation or the differences in the features of the German and English languages. Table~\ref{tab:error-cause} presents full-text examples from the original and translated data. For a  detailed description of the possible issues see Section~\ref{sec:linguistic-analysis}.

\end{document}